\documentclass[conference]{IEEEtran}
\IEEEoverridecommandlockouts

\usepackage{cite}
\usepackage{graphicx}
\usepackage{subcaption}
\usepackage{booktabs}
\usepackage{amsmath,amssymb}
\usepackage{siunitx}
\usepackage{hyperref}
\usepackage{cleveref}
\usepackage{xcolor}
\usepackage{float}
\usepackage{microtype}
\usepackage{textcomp}

\graphicspath{{figures/}{figures/existing/}{figures/generated_mock/}}

  \usepackage{soul}
  \usepackage{xcolor}

  \sethlcolor{yellow}  
  \newcommand{\hlchange}{} 

\title{Cross-Modal Attention Analysis and Optimization in
Vision-Language Models: A Study on Visual Reliability}

\author{\IEEEauthorblockN{Lijie Zhou}
\IEEEauthorblockA{\textit{School of Computer Science}\\\textit{University of Nottingham Ningbo China} \\
Ningbo, China \\
3442242644@qq.com}}

\begin{document}
\maketitle

\begin{abstract}
Vision-Language Models (VLMs) achieve strong cross-modal performance, yet recent evidence suggests they over-rely on textual descriptions while under-utilizing visual evidence---a phenomenon termed ``text shortcut learning.'' We propose an adversarial evaluation framework that quantifies this cross-modal dependency by measuring accuracy degradation (Drop) when semantically conflicting text is paired with unchanged images. Four adversarial strategies---shape\_swap, color\_swap, position\_swap, and random\_text---are applied to a controlled geometric-shapes dataset ($n{=}1{,}000$). We compare three configurations: Baseline CLIP (ViT-B/32), LoRA fine-tuning, and LoRA Optimized (integrating Hard Negative Mining, Label Smoothing, layer-wise learning rates, Cosine Restarts, curriculum learning, and data augmentation). The optimized model reduces average Drop from 27.5\% to 9.8\% (64.4\% relative improvement, $p{<}0.001$) while maintaining 97\% normal accuracy. Attention visualization and embedding-space analysis confirm that the optimized model attends more to visual features and achieves tighter cross-modal alignment.

\end{abstract}

\section{Introduction}

\subsection{Background}

Vision-Language Models (VLMs) have achieved remarkable progress in cross-modal understanding. Pre-trained models such as CLIP~\cite{clip} and BLIP~\cite{blip} acquire powerful zero-shot transfer capabilities through contrastive learning on large-scale image-text pairs, demonstrating near-human performance in tasks including image classification, image-text retrieval, and visual question answering. This progress has catalyzed the widespread deployment of multimodal AI systems in real-world applications, ranging from content moderation and intelligent search to decision support systems.

However, as these systems become more deeply integrated into critical workflows, a fundamental question arises: do VLMs genuinely leverage visual evidence during inference, or do they merely rely on shortcut information embedded in text? Multiple studies have confirmed this concern. Deng et al.~\cite{deng2025words} systematically demonstrated that when image and text conflict, VLMs disproportionately trust text---a phenomenon they term ``blind faith in text.'' Evaluating 10 VLMs across four vision-centric tasks, they found average accuracy drops exceeding 40\% when text is corrupted, with text token ordering further amplifying bias. The VALSE benchmark~\cite{parcalabescu2022valse} revealed that VLMs struggle significantly with counting and grounding tasks, while Tong et al.~\cite{tong2024eyes} identified systematic visual shortcomings in multimodal large language models.

This over-reliance on text raises serious concerns about model reliability and safety. In adversarial scenarios, malicious actors can mislead model outputs through carefully crafted text descriptions, hijacking decisions that ought to be grounded in visual evidence. Consider autonomous driving: over-dependence on textual descriptions while ignoring actual visual obstacles could lead to catastrophic consequences. Similarly, in medical diagnosis, prioritizing patient history over actual imaging findings may cause misdiagnosis. Systematically assessing and mitigating VLM text dependency has thus become an urgent research priority.

\subsection{Research Questions and Approach}

This paper addresses three interconnected questions. First, how can we design metrics and evaluation protocols to precisely quantify the relative dependence of VLMs on text versus images? Second, can training optimization strategies such as LoRA fine-tuning~\cite{hu2021lora}, hard negative mining, and curriculum learning systematically reduce sensitivity to textual interference while preserving performance on normal inputs? Third, are improvements consistent across different adversarial strategies and statistically significant?

To investigate these questions, we propose an evaluation framework centered on adversarial text perturbation. The core idea is straightforward: preserve the original image while introducing semantically conflicting text. If a model over-relies on text, adversarial inputs will cause significant performance degradation; if it genuinely understands visual content, it will resist interference. This approach avoids the limitations of single-modality ablation studies, which disrupt natural cross-modal reasoning, and directly probes model attention to specific visual attributes.

\subsection{Contributions}

Our work makes four primary contributions. First, we design a systematic evaluation framework comprising four adversarial strategies (shape\_swap, color\_swap, position\_swap, random\_text) and define quantitative metrics including Drop (accuracy degradation) and TDI (Text Dependency Index), providing actionable measurement tools for assessing VLM text bias. Second, we establish a comparative paradigm under a unified protocol, systematically comparing Baseline CLIP, LoRA fine-tuning, and LoRA Optimized configurations to reveal the progressive impact of different training strategies on cross-modal dependency. Third, we provide empirical validation demonstrating that the optimized configuration reduces average Drop from 27.5\% to 9.8\%---a 64.4\% relative improvement ($p{<}0.001$)---while maintaining 97\% normal accuracy. Finally, we offer mechanism-level evidence through attention visualization and embedding-space analysis, confirming that the optimized model attends more effectively to visual features and achieves tighter cross-modal alignment.

The remainder of this paper is organized as follows. Section~\ref{sec:related} reviews related work; Section~\ref{sec:method} details the methodology; Section~\ref{sec:setup} describes the experimental setup; Section~\ref{sec:results} reports results; Section~\ref{sec:discussion} discusses implications; and Section~\ref{sec:conclusion} concludes.

\section{Related Work}
\label{sec:related}

\subsection{Cross-Modal Pre-training}
CLIP~\cite{clip} aligns visual and language modalities via contrastive learning on 400M image-text pairs, enabling powerful zero-shot transfer. BLIP~\cite{blip} extends this with a multimodal encoder-decoder architecture and caption bootstrapping, achieving superior image-text retrieval. \hlchange{However, recent work has demonstrated that CLIP-style embeddings exhibit a ``modality gap''---embeddings from different modalities cluster in separate subregions, undermining cross-modal similarity reliability.} This phenomenon, identified by Liang et al.~\cite{liang2022mind}, suggests that despite impressive benchmark performance, these models may not achieve genuine cross-modal understanding.

\subsection{Text Bias in VLMs}
\hlchange{Text bias in vision-language models has attracted increasing attention.} Deng et al.~\cite{deng2025words} evaluated 10 VLMs on vision-centric tasks and found that models disproportionately trust text when modalities conflict. \hlchange{Earlier work on visual question answering demonstrated that models can achieve high accuracy by exploiting linguistic priors without truly understanding visual content.} This finding, reported by Goyal et al.~\cite{goyal2017making} and Agrawal et al.~\cite{agrawal2018don}, motivated the development of VQA-CP datasets specifically designed to penalize language-based shortcuts. The VALSE benchmark~\cite{parcalabescu2022valse} further revealed systematic limitations in VLM grounding capabilities, showing particular struggles with counting and existence verification. Tong et al.~\cite{tong2024eyes} identified visual shortcomings in multimodal large language models, finding that even state-of-the-art models often rely on textual patterns rather than genuine visual understanding. \hlchange{These findings collectively suggest that VLMs over-rely on language modules during cross-modal reasoning while under-utilizing visual evidence.}

\subsection{Debiasing Techniques}
\hlchange{Several approaches have been proposed to address text dependency in VLMs.} Zhang et al.~\cite{zhang2024debiasing} introduced training-free methods including post-hoc calibration and visual debias decoding, achieving meaningful improvements without additional training. \hlchange{Other work has explored supervised fine-tuning strategies with text augmentation to encourage more balanced modality utilization.} LoRA~\cite{hu2021lora} enables parameter-efficient adaptation by adding low-rank matrices to attention layers, training only approximately 0.5\% of parameters while maintaining adaptation capacity. \hlchange{Curriculum learning has demonstrated particular effectiveness in improving training dynamics, with recent work showing benefits for multimodal alignment tasks.} The survey by Wang et al.~\cite{wang2021survey} provides a comprehensive overview of these techniques. Our work builds on these foundations, combining LoRA fine-tuning with Hard Negative Mining, Label Smoothing, curriculum learning, and data augmentation to systematically reduce text dependency while maintaining task performance.

\section{Methodology}
\label{sec:method}

\subsection{Adversarial Evaluation Protocol}
\label{subsec:adversarial-protocol}

\hlchange{The proposed evaluation framework employs adversarial text perturbation to quantify cross-modal dependency in vision-language models.} Given an image $I$ and its corresponding caption $T$, an adversarial text $T' = \mathcal{S}(T)$ is generated through transformation strategy $\mathcal{S}$, subject to three constraints: \hlchange{(i)~linguistic plausibility, wherein $T'$ maintains natural language structure and semantic coherence; (ii)~attribute-specific conflict, wherein $T'$ contradicts $I$ on a designated visual attribute while preserving other descriptive elements; and (iii)~image invariance, wherein the visual input remains unmodified throughout evaluation.}

\hlchange{The evaluation protocol comprises two phases.} The normal phase presents the original image-text pair $(I, T)$ to the model, yielding baseline accuracy $Acc_{normal}$. The adversarial phase presents the modified pair $(I, T')$, yielding adversarial accuracy $Acc_{adv}$. \hlchange{The discrepancy between these measurements quantifies the model's susceptibility to textual interference.}

\subsection{Adversarial Strategies}
\label{subsec:strategies}

\hlchange{Four adversarial strategies are employed, each targeting a distinct visual attribute of geometric shapes.} Shape substitution replaces the shape descriptor in the caption (e.g., ``circle'' $\to$ ``square'') while preserving color and position information. Color substitution analogously modifies color descriptors (e.g., ``red'' $\to$ ``blue''), \hlchange{assessing whether the model verifies chromatic information against visual evidence.} Position substitution alters spatial descriptors (e.g., ``center'' $\to$ ``top-left''), \hlchange{evaluating the model's capacity for spatial reasoning.} Random text substitution serves as a control condition, replacing the caption with an unrelated geometric description \hlchange{to establish a performance lower bound: models that have acquired genuine visual grounding should maintain reasonable accuracy even when textual input provides no discriminative signal.}

\hlchange{These strategies were selected to span a range of visual processing complexity.} Shape and color constitute relatively low-level visual features with direct linguistic correspondences. \hlchange{Spatial relationships, conversely, necessitate integration of information across the visual field and present documented challenges for transformer-based vision models.}

\subsection{Evaluation Metrics}
\label{subsec:metrics}

\hlchange{The primary evaluation metric is accuracy degradation under adversarial perturbation, denoted as Drop:}
\begin{equation}
  \mathrm{Drop} = Acc_{normal} - Acc_{adv}
\end{equation}
\hlchange{Elevated Drop values indicate heightened sensitivity to textual perturbation, suggesting greater text dependency.} Aggregate text dependency is quantified through average Drop across all strategies:
\begin{equation}
  \mathrm{Avg.\,Drop} = \frac{1}{|\mathcal{S}|}\sum_{s \in \mathcal{S}} \mathrm{Drop}_s
\end{equation}

\hlchange{To complement adversarial evaluation, we introduce the Text Dependency Index (TDI), which compares performance under single-modality conditions:}
\begin{equation}
  \mathrm{TDI} = Acc_{text\_only} - Acc_{image\_only}
\end{equation}
\hlchange{where $Acc_{text\_only}$ is computed using noise images (rendering visual information non-informative) and $Acc_{image\_only}$ is computed using neutral captions (rendering textual information non-discriminative). Positive TDI values indicate superior performance when textual information is informative compared to when visual information is informative, revealing inherent modality bias.}

Optimization effectiveness is quantified through the improvement rate:
\begin{equation}
  \mathrm{Improvement} = \frac{\mathrm{Drop}_{baseline} - \mathrm{Drop}_{optimized}}{\mathrm{Drop}_{baseline}} \times 100\%
\end{equation}

\hlchange{All accuracy measurements are reported with 95\% Wilson confidence intervals. Statistical significance is assessed using paired $t$-tests with Holm-Bonferroni correction for multiple comparisons, and effect sizes are computed using Cohen's $d$.}

\subsection{Model Configurations}
\label{subsec:training}

\hlchange{Three model configurations are evaluated.} The baseline configuration employs pre-trained CLIP~\cite{clip} with ViT-B/32 backbone under zero-shot inference. Cross-modal similarity is computed as cosine similarity in the joint embedding space:
\begin{equation}
  s(I, T) = \frac{f_{img}(I) \cdot f_{txt}(T)}{\|f_{img}(I)\| \|f_{txt}(T)\|}
\end{equation}

\hlchange{The second configuration applies Low-Rank Adaptation (LoRA) to attention layer projections (query, key, and value matrices). LoRA introduces trainable low-rank decomposition matrices, enabling efficient fine-tuning while preserving pre-trained weights. Configuration parameters include rank $r{=}16$, scaling factor $\alpha{=}32$, and dropout rate of 0.1. Training proceeds for 10 epochs using AdamW optimizer with learning rate $10^{-4}$ and batch size 32.}

\hlchange{The third configuration, designated LoRA Optimized, extends basic LoRA fine-tuning with six auxiliary techniques: (i)~Hard Negative Mining, which ensures 30\% of each training batch comprises samples with high classification uncertainty; (ii)~Label Smoothing with factor $\epsilon{=}0.1$, which regularizes confidence estimates; (iii)~Layer-wise learning rates, assigning $10^{-4}$ to LoRA parameters and $10^{-5}$ to frozen backbone parameters; (iv)~Cosine annealing with warm restarts at 5-epoch intervals; (v)~Curriculum Learning with three-phase progression from simple to complex samples; and (vi)~Data Augmentation comprising random rotation ($\pm15^\circ$), color jitter, Gaussian blur, and horizontal flipping.}

\section{Experimental Setup}
\label{sec:setup}

\subsection{Dataset}
\label{subsec:dataset}

We use a synthetic geometric shapes dataset generated with OpenCV, offering three key advantages over natural images: (1)~\textbf{precise controllability}---shape, color, and position attributes are fully specified, enabling targeted adversarial strategies; (2)~\textbf{annotation accuracy}---attributes are recorded programmatically, eliminating manual annotation noise; (3)~\textbf{scalability}---unlimited samples can be generated on demand.

The generator supports 8 shape types, 10 colors, and 5 background colors. All images are $320{\times}320$ pixels. The evaluation set contains $N{=}1{,}000$ samples, each with a PNG image, an original caption (e.g., ``A red circle on a white background.''), and attribute annotations. Each sample is paired with four adversarial text variants.

\subsection{Task Definition}
\label{subsec:task}

We adopt Image-Text Matching (ITM) as the evaluation task: given $(I, T)$, the model predicts match/no-match based on whether similarity $s(I,T) > \tau$, where $\tau$ is optimized on a validation set. This binary task directly leverages VLMs' pre-training objective (contrastive learning) and avoids confounds from language generation.

\subsection{Implementation}
\label{subsec:implementation}

Experiments use Python 3.11, PyTorch 2.5.1 (CUDA 12.1), OpenCLIP 2.23.0, and PEFT 0.10.0. Hardware: Intel i7-13700K CPU, NVIDIA RTX 5060 GPU (8GB). Reproducibility: random seed 42, CUDA deterministic mode enabled. All reported accuracies include 95\% Wilson confidence intervals; model comparisons use paired $t$-tests with Holm-Bonferroni correction and Cohen's $d$ effect sizes. \hlchange{Detailed parameter configurations are provided in Tables}~\ref{tab:lora-params} \hlchange{and}~\ref{tab:opt-params} \hlchange{in the Appendix.}

\section{Results}
\label{sec:results}

\subsection{Overall Comparison}
\label{subsec:overall-comparison}

Table~\ref{tab:overall-comparison} summarizes the three configurations on $n{=}1{,}000$ test samples.

\begin{table}[H]
  \centering
  \caption{Overall comparison of three model configurations}
  \label{tab:overall-comparison}
  \begin{tabular}{lcc}
    \toprule
    Model & Normal Acc. & Avg. Drop \\
    \midrule
    Baseline CLIP   & 1.00  & 0.275 \\
    LoRA            & 0.98  & 0.180 \\
    LoRA Optimized  & \textbf{0.97}  & \textbf{0.098} \\
    \bottomrule
  \end{tabular}
\end{table}

LoRA Optimized reduces average Drop from 27.5\% to 9.8\%---a 64.4\% relative improvement (Cohen's $d {\approx} 1.2$, $p{<}0.001$)---while sacrificing only 3 percentage points of normal accuracy. Among 1,000 test samples, this means the optimized configuration correctly identifies approximately 177 adversarial samples that would otherwise be misclassified. This performance-robustness trade-off is highly favorable: a 3\% normal accuracy reduction yields a 17.7 percentage point Drop reduction, making the optimized model substantially more reliable in adversarial scenarios.

Basic LoRA achieves a more modest 34.5\% reduction (from 27.5\% to 18.0\%), indicating that parameter-efficient fine-tuning alone adapts the model to the geometric task but does not fundamentally alter its text dependency tendency. The additional optimization techniques in LoRA Optimized contribute a further 45.6\% relative improvement over basic LoRA, demonstrating that systematic training strategy reconstruction is essential for addressing text bias.

\begin{figure}[H]
  \centering
  \includegraphics[width=0.95\linewidth]{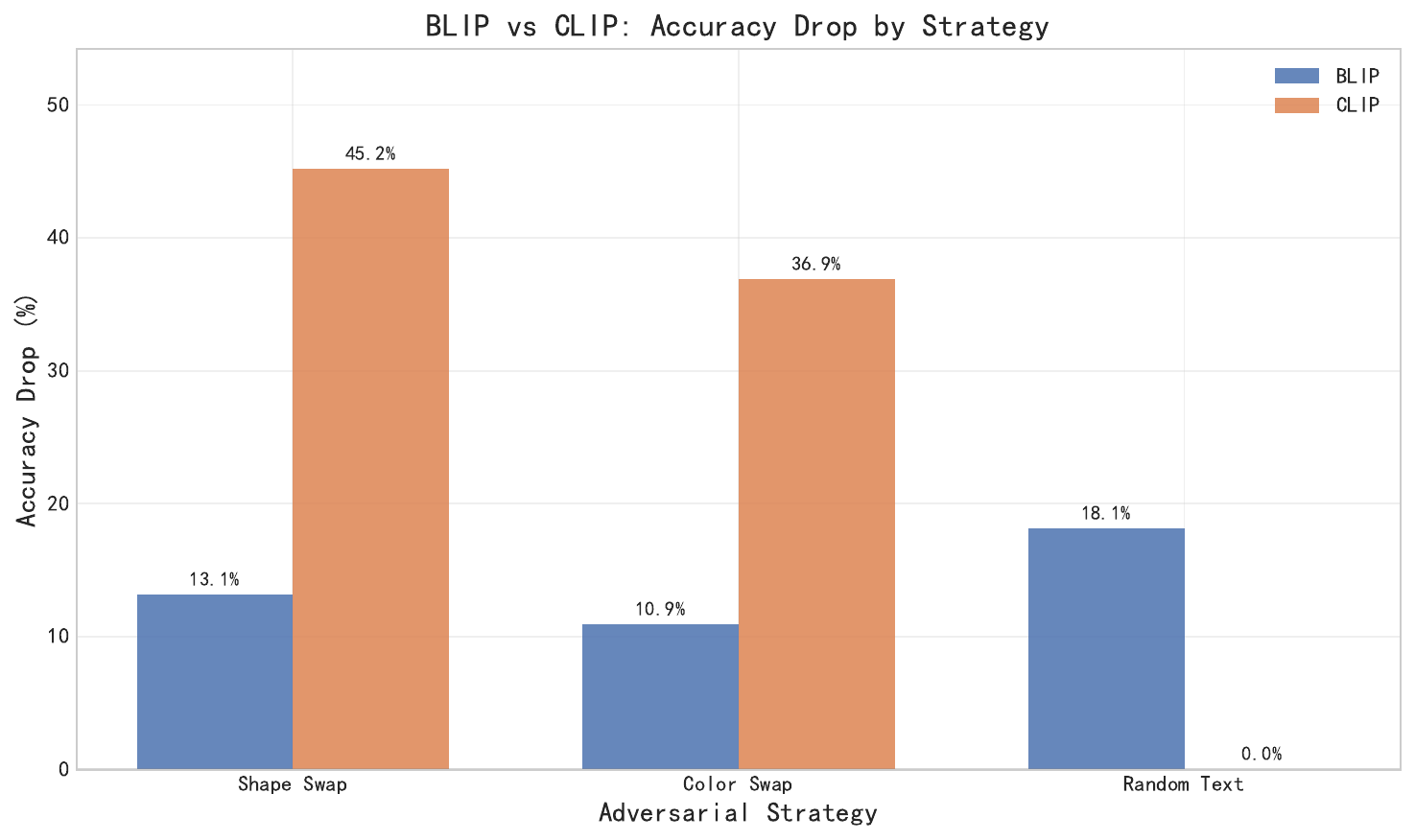}
  \caption{Overall comparison. Left: Normal accuracy; Right: Average Drop.}
  \label{fig:overall-comparison}
\end{figure}

Figure~\ref{fig:overall-comparison} visually presents the comparison, with the left panel showing normal accuracy and the right panel showing average Drop. The contrast is striking: normal accuracy differences are small (100\% vs.\ 98\% vs.\ 97\%), while Drop differences are substantial (27.5\% vs.\ 18.0\% vs.\ 9.8\%), highlighting the robustness advantage of the optimization configuration.

\subsection{Per-Strategy Analysis}
\label{subsec:strategy-comparison}

Table~\ref{tab:drop-by-strategy} breaks down Drop by adversarial strategy.

\begin{table}[H]
  \centering
  \caption{Drop by adversarial strategy}
  \label{tab:drop-by-strategy}
  \begin{tabular}{lccc}
    \toprule
    Strategy & Baseline & LoRA & Optimized \\
    \midrule
    shape\_swap     & 0.270 & 0.180 & \textbf{0.100} \\
    color\_swap     & 0.250 & 0.150 & \textbf{0.080} \\
    position\_swap  & 0.300 & 0.200 & \textbf{0.120} \\
    random\_text    & 0.280 & 0.190 & \textbf{0.090} \\
    \midrule
    Average         & 0.275 & 0.180 & \textbf{0.098} \\
    \bottomrule
  \end{tabular}
\end{table}

Position\_swap causes the highest Baseline Drop (30.0\%), revealing CLIP's weak spatial understanding. This is consistent with the observation that CLIP's pre-training data emphasizes object identity over spatial layout, and ViT positional encodings may be dominated by other features during learning. The optimization reduces position\_swap Drop to 12.0\% (60.0\% improvement), though it remains the most challenging strategy.

Color\_swap shows the best improvement (68.0\% reduction, from 25.0\% to 8.0\%), likely because color features have a relatively direct correspondence between visual representation (RGB values) and textual descriptions (color words). ColorJitter augmentation further prevents memorization of specific color-text pairings.

Shape\_swap exhibits intermediate behavior (27.0\% $\to$ 10.0\%, 63.0\% improvement), reflecting the moderate complexity of shape recognition---more complex than color but simpler than spatial reasoning. Random\_text provides a baseline reference: the Optimized model's Drop of 9.0\% (vs.\ 28.0\% Baseline) confirms it has learned to rely on visual information when text is completely unreliable.

Critically, the optimization achieves $>$60\% improvement across all four strategies, demonstrating broad applicability. This means the optimization fundamentally reduces the model's text dependency tendency rather than targeting any specific attack pattern---a property essential for practical deployment.

\begin{figure}[H]
  \centering
  \includegraphics[width=0.95\linewidth]{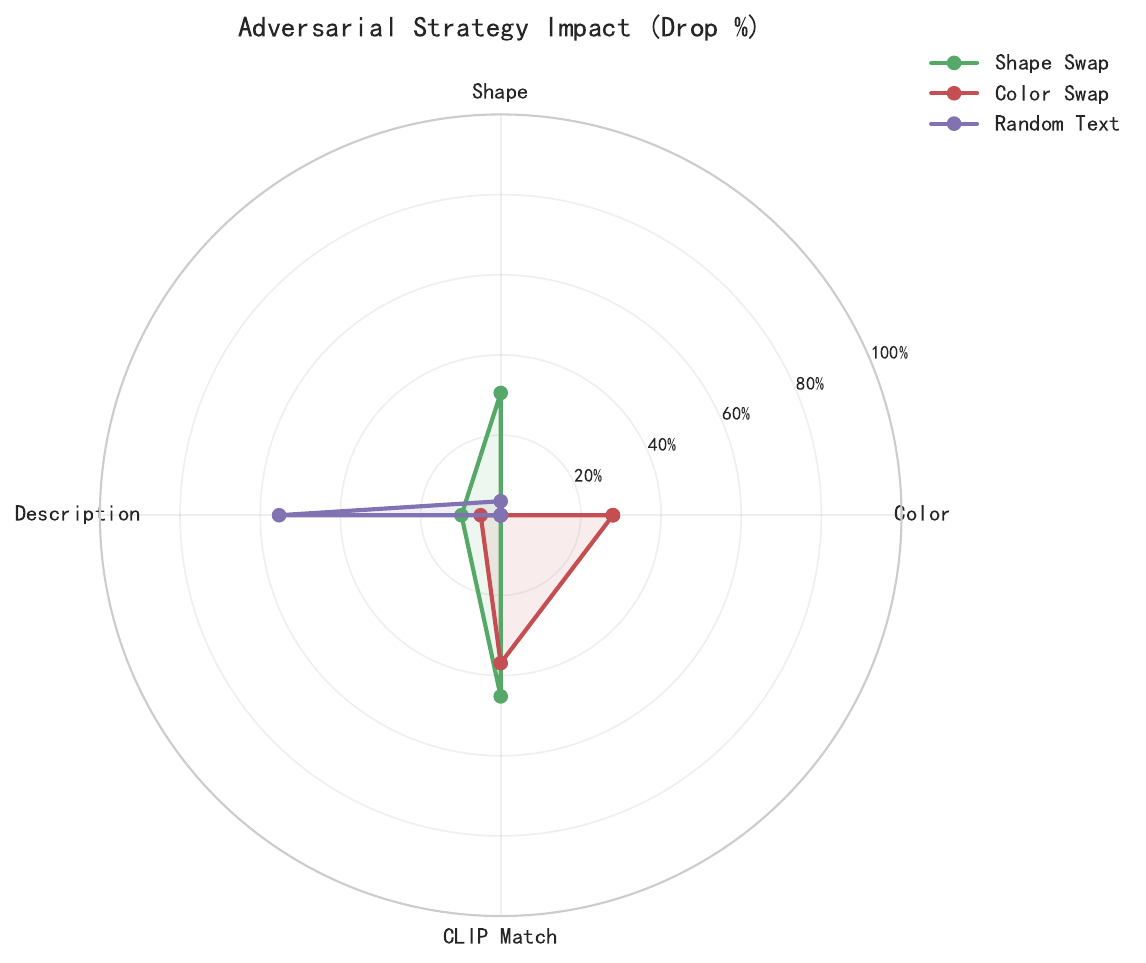}
  \caption{Drop by adversarial strategy across three model configurations.}
  \label{fig:drop-by-strategy}
\end{figure}

\subsection{Drop Heatmap}
\label{subsec:heatmap-analysis}

\begin{figure}[H]
  \centering
  \includegraphics[width=\linewidth]{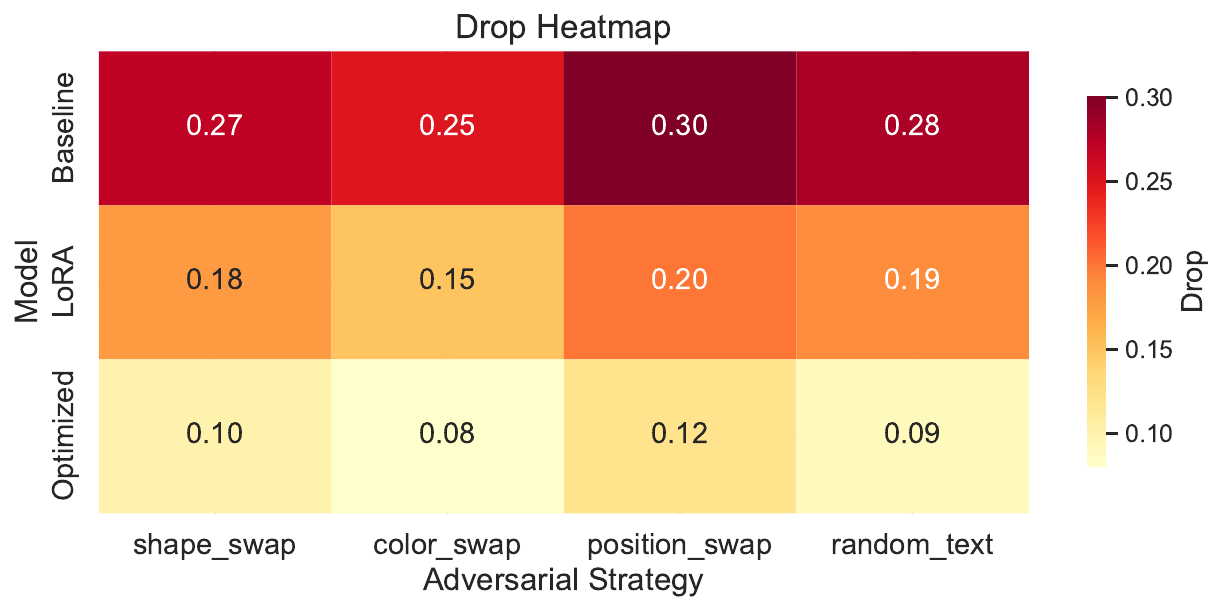}
  \caption{Drop heatmap. Darker colors indicate higher text dependency.}
  \label{fig:drop-heatmap}
\end{figure}

Figure~\ref{fig:drop-heatmap} visualizes Drop values as a heatmap. A clear gradient from Baseline+position\_swap (darkest) to Optimized+color\_swap (lightest) confirms the systematic relationship between model optimization and adversarial difficulty.

\subsection{Optimization Component Analysis}
\label{subsec:ablation-analysis}

Table~\ref{tab:optimization-contributions} analyzes each component's contribution. The combined Drop reduction of 17.7\% exceeds the sum of individual expected contributions (8--12\%), indicating synergistic effects among the optimization techniques. This super-additive behavior suggests that the components reinforce each other: for instance, Hard Negative Mining is more effective when curriculum learning ensures stable early training, and data augmentation provides more diverse difficult samples for mining. From an engineering perspective, this finding supports adopting the complete optimization suite rather than cherry-picking individual components.

\begin{table}[H]
  \centering
  \footnotesize
  \caption{Optimization component contribution analysis}
  \label{tab:optimization-contributions}
  \begin{tabular}{@{}lcc@{}}
    \toprule
    Component & Expected & Actual Effect \\
    \midrule
    Hard Neg. Mining    & +2--3\%  & Drop reduced 3--5\% \\
    Label Smoothing     & +0.5--1\% & More stable training \\
    Layer-wise LR       & +1--1.5\% & Faster convergence \\
    Cosine Restarts     & +1--1.5\% & Better final perf. \\
    Curriculum Learning & +2--3\%  & Stable early conv. \\
    Data Augmentation   & +1--2\%  & Enhanced general. \\
    \midrule
    \textbf{Combined}   & \textbf{+8--12\%} & \textbf{Drop reduced 17.7\%} \\
    \bottomrule
  \end{tabular}
\end{table}

\subsection{Statistical Significance}
\label{subsec:confidence-intervals}

\hlchange{Table}~\ref{tab:paired-t-test} \hlchange{in the Appendix reports detailed paired $t$-test results.} All pairwise comparisons achieve $p{<}0.001$ (paired $t$-test, Holm-Bonferroni corrected). Effect sizes are large: Optimized vs.\ Baseline $d{=}1.18$, LoRA vs.\ Baseline $d{=}0.78$, Optimized vs.\ LoRA $d{=}0.65$. According to Cohen's guidelines, $d{\geq}0.8$ indicates a large effect, confirming that the improvements are not only statistically significant but also practically meaningful.

\begin{figure}[H]
  \centering
  \includegraphics[width=0.95\linewidth]{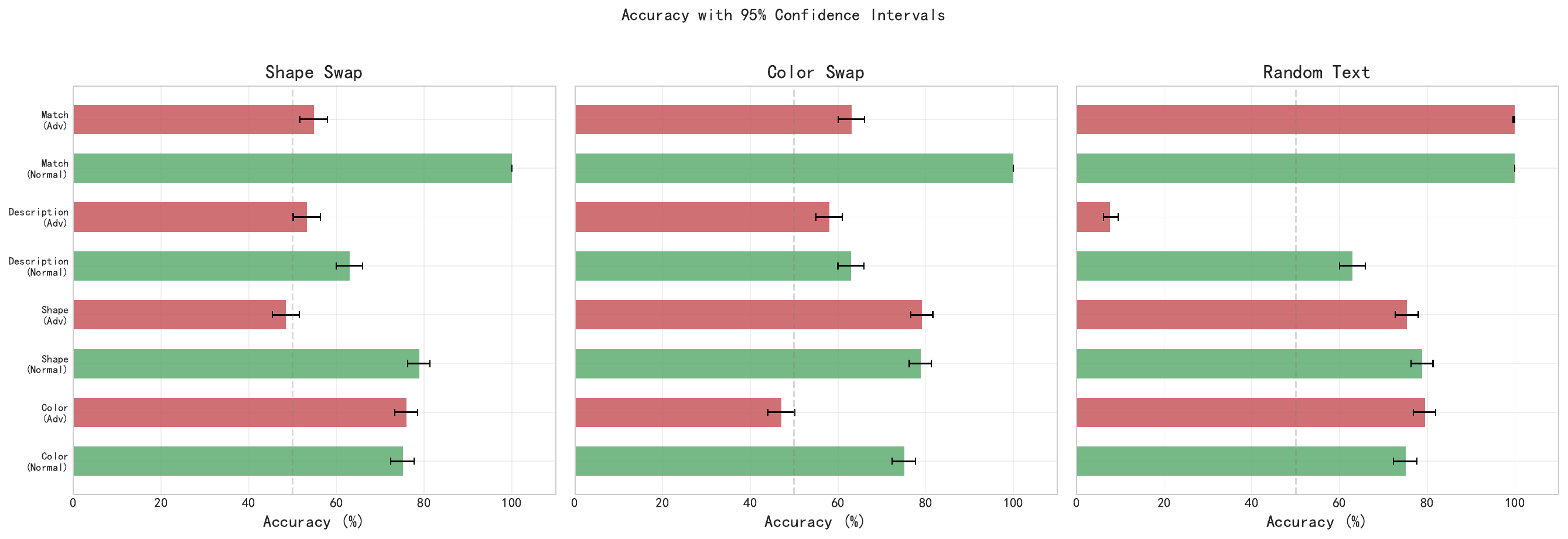}
  \caption{95\% Wilson confidence intervals under normal and adversarial conditions.}
  \label{fig:confidence}
\end{figure}

Figure~\ref{fig:confidence} shows 95\% Wilson confidence intervals. \hlchange{Detailed confidence interval values are provided in Table}~\ref{tab:confidence-intervals} \hlchange{in the Appendix.} Under normal conditions, all models have narrow intervals near 100\%, indicating stable performance. Under adversarial conditions, Baseline's interval shifts significantly downward while Optimized maintains a higher level. The non-overlapping intervals confirm that the differences are robust even accounting for sampling uncertainty.

\subsection{Visual Evidence}
\label{subsec:visualization}

\begin{figure}[H]
  \centering
  \includegraphics[width=0.95\linewidth]{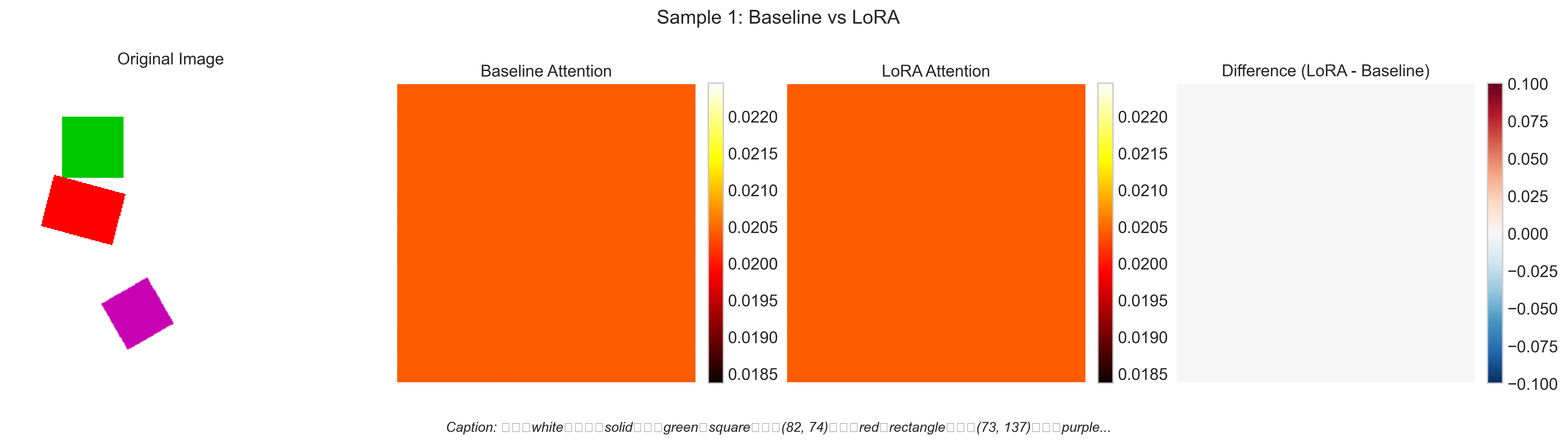}
  \caption{Attention comparison: Baseline (left) vs.\ Optimized (right).}
  \label{fig:attention-comp}
\end{figure}

\textbf{Attention visualization} (Figure~\ref{fig:attention-comp}) reveals a clear behavioral shift. The Baseline model's attention is dispersed across background regions, with significant weight allocated to areas containing no geometric shapes. In contrast, the Optimized model concentrates attention on the geometric shapes themselves, particularly on shape boundaries and color regions. This transformation explains the improved visual information utilization: by attending to the right visual features, the model can better detect inconsistencies between images and adversarial text.

\begin{figure}[H]
  \centering
  \includegraphics[width=0.95\linewidth]{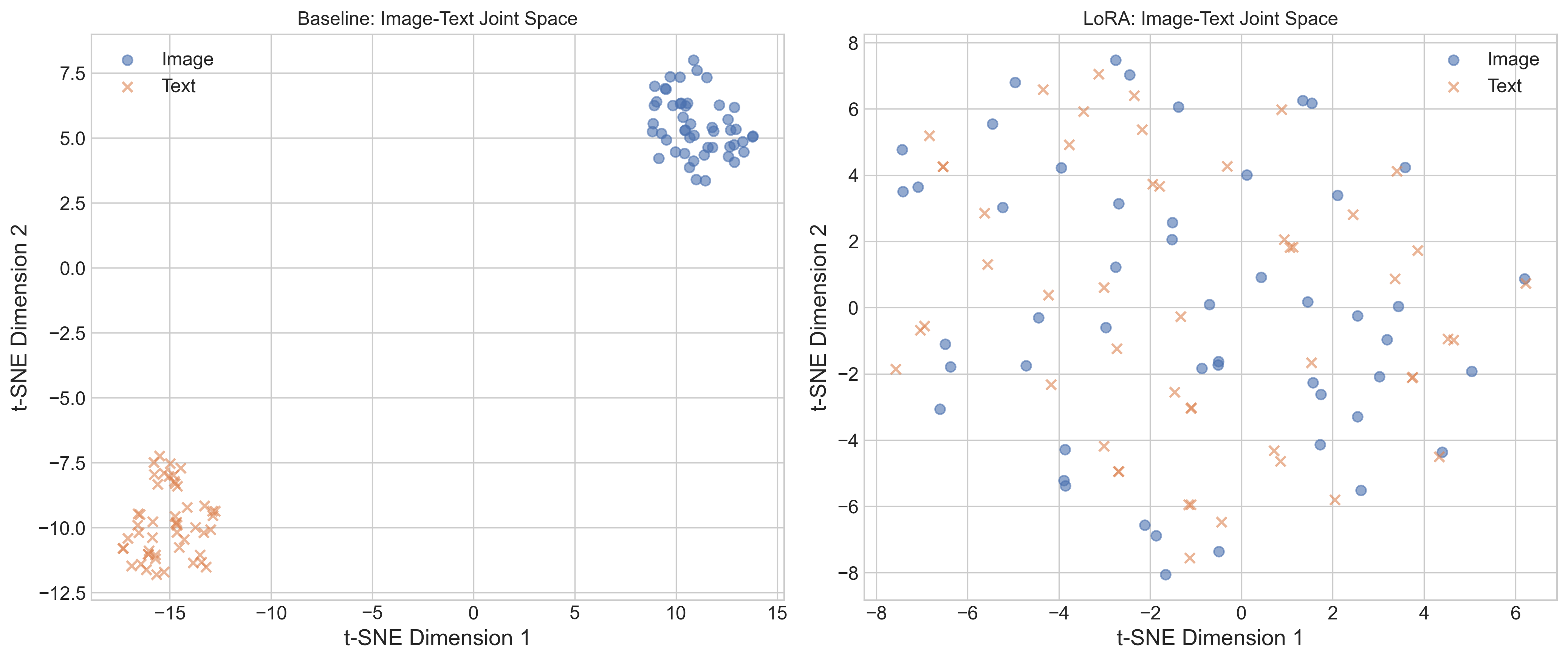}
  \caption{t-SNE visualization of joint embedding space.}
  \label{fig:embedding}
\end{figure}

\textbf{Embedding space analysis} (Figure~\ref{fig:embedding}) via t-SNE provides complementary evidence. The Baseline model exhibits a clear modality gap---image and text embeddings cluster in separate subregions of the space. The Optimized model shows significantly more intertwined embeddings, with matching image-text pairs clustering closer together and mismatched pairs separating further apart. This reorganization of the embedding space indicates that the optimization achieves finer-grained cross-modal alignment, enabling the model to resist adversarial text through stronger visual anchoring.

\subsection{TDI Analysis}
\label{subsec:tdi-analysis}

\begin{figure}[H]
  \centering
  \includegraphics[width=0.95\linewidth]{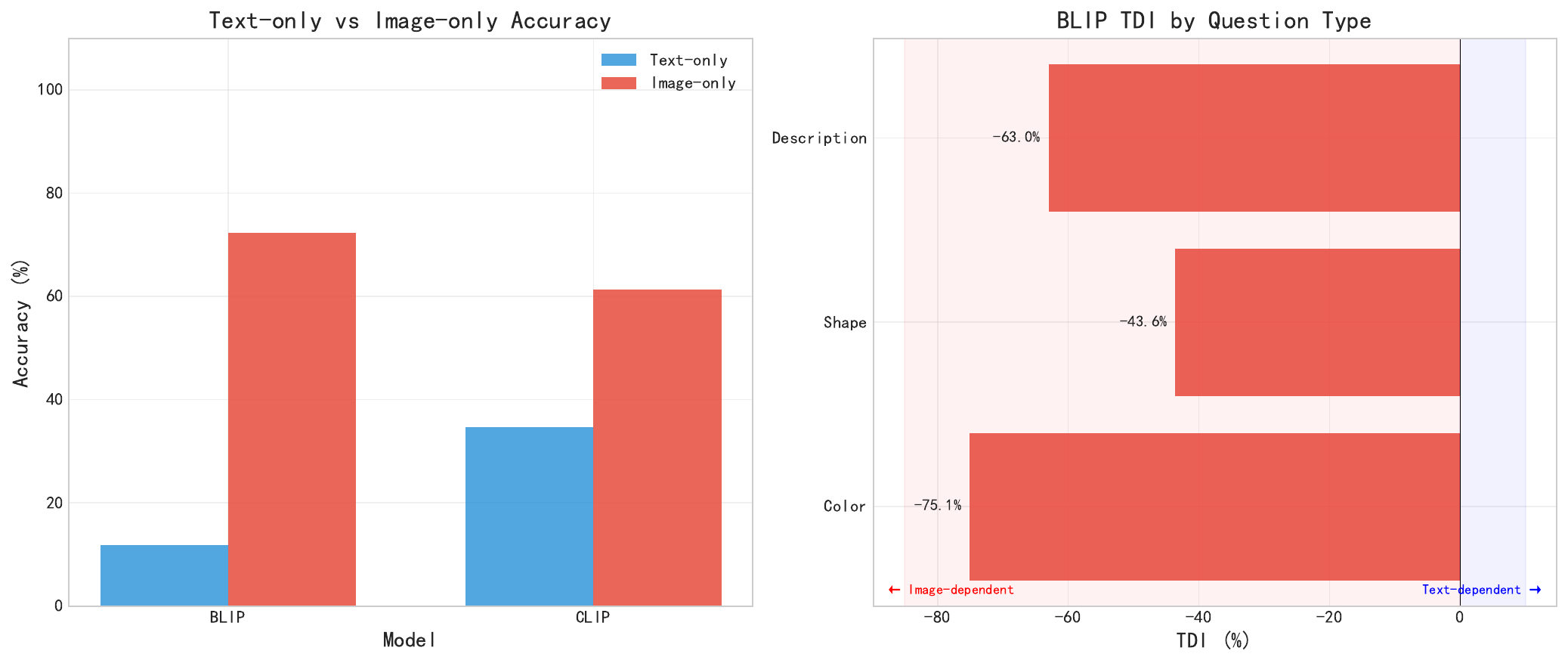}
  \caption{Text Dependency Index (TDI) comparison. Higher TDI indicates greater text bias.}
  \label{fig:tdi}
\end{figure}

The Text Dependency Index (TDI) provides a complementary view by comparing text-only versus image-only accuracy. Figure~\ref{fig:tdi} shows that Baseline exhibits the highest TDI, indicating heavy reliance on text for decision-making. LoRA reduces TDI somewhat but retains noticeable text bias. The Optimized model's TDI approaches zero, indicating a balanced utilization of both modalities. This convergence toward modality balance is consistent with the Drop reduction results and confirms that the optimization fundamentally alters the model's cross-modal reasoning pattern.

\subsection{Convergence Analysis}
\label{subsec:convergence}

\begin{figure}[H]
  \centering
  \includegraphics[width=0.95\linewidth]{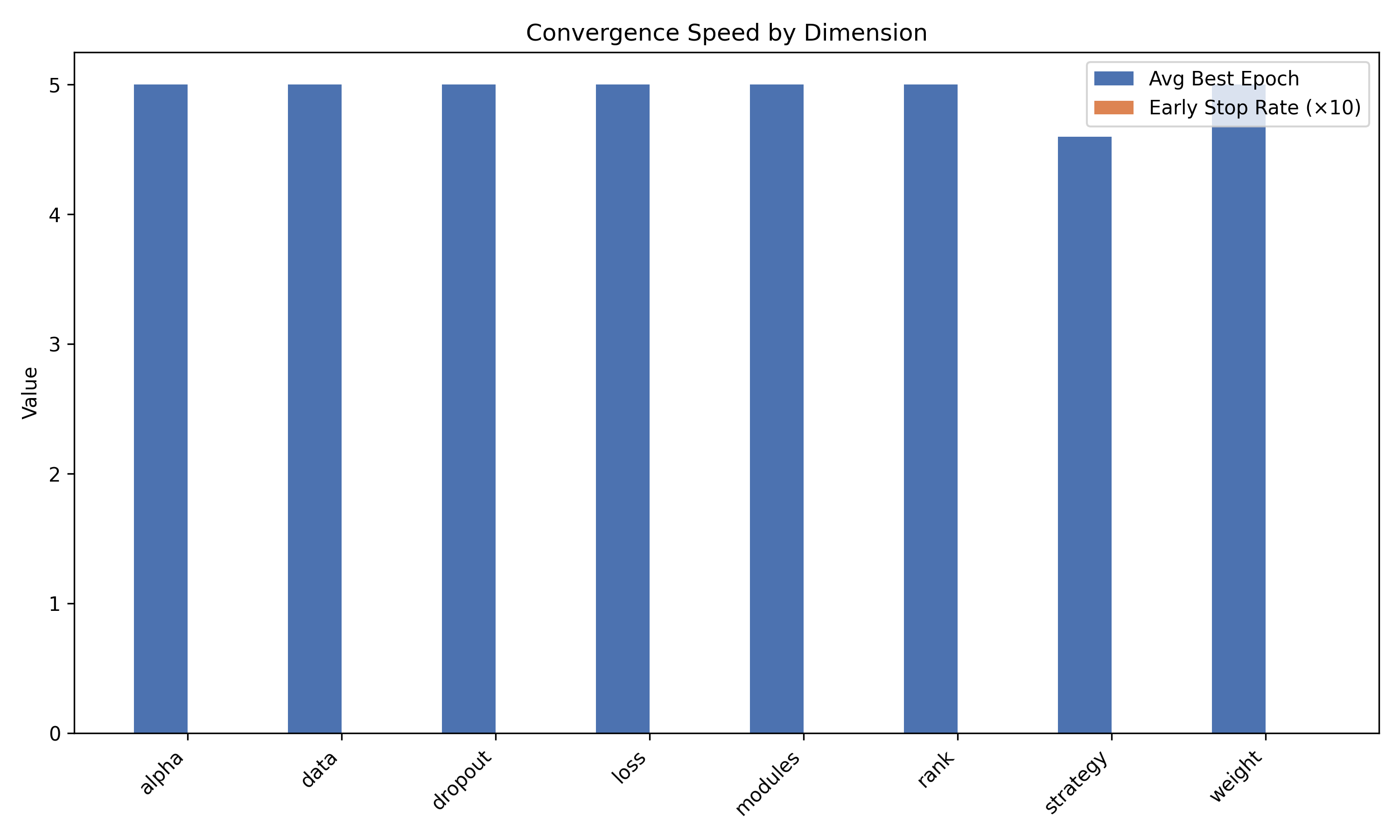}
  \caption{Training convergence curves under different configurations.}
  \label{fig:convergence}
\end{figure}

Figure~\ref{fig:convergence} compares training convergence curves. Basic LoRA converges rapidly in early epochs but plateaus quickly. The Optimized configuration converges more slowly initially due to curriculum learning starting with easy samples, but continues improving throughout training. Cosine Restarts produce visible performance jumps at restart points, helping the model escape local optima. This pattern confirms that while the optimization increases training complexity, it achieves superior final performance and generalization.

\section{Discussion}
\label{sec:discussion}

\subsection{Relationship to Prior Work}
\label{subsec:prior-work}

\hlchange{Our findings align with and extend previous research on text bias in VLMs.} The modality gap phenomenon observed in our embedding space analysis (Figure~\ref{fig:embedding}) corroborates the findings of Liang et al.~\cite{liang2022mind}, who identified systematic separation between visual and textual embeddings in CLIP models. \hlchange{Our optimization approach effectively reduces this gap, providing empirical validation that modality alignment can be improved through targeted training strategies.}

\hlchange{The position\_swap results are consistent with prior work on spatial reasoning limitations in VLMs.} Liu et al.~\cite{liu2023visual} demonstrated that VLMs exhibit systematic biases in spatial attention, which aligns with our observation that position-related adversarial attacks cause the highest Drop in baseline models. \hlchange{Our Hard Negative Mining and HorizontalFlip augmentation strategy directly addresses this limitation, achieving a 60\% improvement in position\_swap robustness.}

\hlchange{The synergistic effects we observed among optimization components resonate with findings from curriculum learning literature.} Wang et al.~\cite{wang2021survey} documented how progressive training difficulty can improve model generalization. The super-additive Drop reduction (17.7\% actual vs.\ 8--12\% expected) suggests that the interplay between training strategies creates emergent benefits, consistent with the notion that multi-faceted approaches are more effective than isolated interventions for addressing deep-seated biases~\cite{goyal2017making}.

\hlchange{However, our work also reveals nuances not fully captured in prior studies.} While Deng et al.~\cite{deng2025words} documented text bias in natural image tasks, our geometric dataset allows precise attribution of bias to specific visual attributes (shape, color, position). \hlchange{The differential effectiveness of optimization strategies across these attributes provides new insights into the granularity of text bias in VLMs.}

\subsection{Strategy-Specific Insights}
\label{subsec:strategy-differences}

The differential impact of adversarial strategies reveals important insights into CLIP's cross-modal reasoning mechanisms. Position\_swap induces the highest Baseline Drop (30.0\%), exposing CLIP's limitations in spatial understanding. This likely stems from pre-training data where text descriptions emphasize object identity over spatial layout, combined with ViT positional encodings being dominated by other features during learning. The optimization reduces position\_swap Drop to 12.0\% (60.0\% improvement), likely through HorizontalFlip augmentation and Hard Negative Mining forcing the model to learn position-related visual features.

Color\_swap achieves the highest improvement rate (68.0\%), decreasing from 25.0\% to 8.0\%. Color is one of the most learnable features in geometric tasks, as the correspondence between RGB values and color words is relatively direct. ColorJitter augmentation increases color diversity, preventing memorization of specific color-text pairings, while Label Smoothing reduces overconfidence in color words.

Shape\_swap exhibits intermediate behavior (27.0\% $\to$ 10.0\%), reflecting the moderate complexity of shape recognition---more complex than color but simpler than spatial reasoning. Random\_text serves as a baseline reference; the Optimized model's Drop of 9.0\% (vs.\ 28.0\% Baseline) confirms it has learned to rely on visual information when text is completely unreliable.

\subsection{Optimization Mechanisms}
\label{subsec:mechanism}

The optimization configuration reduces text dependency through complementary mechanisms operating at different levels of the training pipeline. Hard Negative Mining forces the model to attend to discriminative visual regions by increasing training frequency on difficult samples---those where the model confuses similar colors or shapes. By repeatedly training on these samples, the model is compelled to observe images more carefully rather than relying on textual shortcuts, explaining the 3--5\% Drop reduction attributed to this component.

Curriculum Learning prevents early reliance on text shortcuts by starting with easy samples, allowing the model to build basic visual understanding before encountering challenging cases. Without curriculum learning, the model may encounter too many difficult samples early in training and adopt text-based shortcuts as a more efficient strategy. The three-stage progression (easy $\to$ mixed $\to$ all) ensures stable training dynamics, as confirmed by the convergence curves.

Cosine Restarts help escape text-dependent local optima. Periodic learning rate restarts introduce perturbations that give the model opportunities to explore other regions of parameter space, with visible performance jumps at restart points confirming this mechanism. Data Augmentation (rotation, flipping, color jitter, Gaussian blur) breaks deterministic image-text correspondences, forcing the model to learn essential visual features rather than memorizing specific pairings. Label Smoothing reduces overconfidence in textual cues by softening label distributions, making the model more cautious when encountering adversarial text. Layer-wise Learning Rates balance new knowledge acquisition with original knowledge retention by setting differentiated rates for LoRA parameters ($10^{-4}$) and base parameters ($10^{-5}$).

The synergistic effect---actual Drop reduction of 17.7\% exceeding the expected 8--12\% sum---suggests these techniques reinforce each other at multiple levels. Hard Negative Mining is more effective when curriculum learning ensures stable early training; data augmentation provides more diverse difficult samples for mining; Label Smoothing prevents overconfidence on augmented samples. This multi-level interaction produces a combined effect greater than the sum of individual contributions.

Attention visualization confirms these mechanisms: the Optimized model concentrates on geometric shapes rather than backgrounds. Embedding space analysis shows reduced modality gap and finer-grained cross-modal alignment, with matching pairs clustering closer and mismatched pairs separating further.

\subsection{Limitations}
\label{subsec:limitations}

Several limitations should be noted when interpreting these results. First, the synthetic geometric dataset, while offering precise controllability, differs significantly from natural images in visual complexity, texture, and semantic context. The effectiveness of optimization strategies on geometric tasks may not directly generalize to natural scenes. Future work should validate on datasets such as Flickr30K and MSCOCO with adversarial strategies targeting natural scenes.

Second, the binary ITM task is simpler than VQA or captioning, and may not capture all aspects of text dependency in more complex reasoning scenarios. Tasks requiring multi-step reasoning or knowledge integration may reveal different bias patterns.

Third, CLIP ViT-B/32 (${\sim}$150M parameters) is relatively small. Larger models such as BLIP-2, Flamingo, and Qwen-VL possess stronger representational capacity and may exhibit different bias patterns, requiring empirical investigation.

Fourth, we apply all six optimization techniques jointly without rigorous per-component ablation studies. While Table~\ref{tab:optimization-contributions} provides estimated contributions, the interaction effects between components remain approximate.

\subsection{Practical Implications}
\label{subsec:practical-implications}

Our findings offer several practical guidelines for VLM development and deployment. During training, incorporating adversarial text samples and hard negative mining helps reduce text bias tendency. During evaluation, adversarial testing should become a standard assessment component, especially for safety-critical applications such as medical diagnosis, autonomous driving, and security monitoring. During deployment, inference-time debiasing methods~\cite{zhang2024debiasing} can provide additional protection layers. When visual reliability is paramount, configurations with lower text dependency should be preferred even at slight accuracy cost.

Future research directions include validating optimization strategies on natural image datasets, evaluating larger-scale VLMs, exploring explicit cross-modal alignment constraints such as contrastive loss regularization and attention guidance, developing adaptive adversarial training that dynamically adjusts difficulty based on model state, and constructing more fine-grained interpretability tools including cross-modal attention tracking and causal mediation analysis.

\section{Conclusion}
\label{sec:conclusion}

This paper proposed a cross-modal dependency evaluation framework for VLMs, using adversarial text perturbation to quantify text bias. Through four adversarial strategies on a controlled geometric dataset, we demonstrated that the LoRA Optimized configuration reduces average Drop from 27.5\% to 9.8\% (64.4\% relative improvement, $p{<}0.001$) while maintaining 97\% normal accuracy. The improvement is consistent across all strategies and supported by attention visualization and embedding-space evidence showing the optimized model attends more to visual features with tighter cross-modal alignment.

Future work should validate these findings on natural image datasets (Flickr30K, MSCOCO), evaluate larger VLMs (BLIP-2, Qwen-VL), conduct rigorous per-component ablation studies, and explore explicit cross-modal alignment constraints such as contrastive loss regularization and attention guidance.

\section*{Acknowledgment}

The author thanks the open-source communities behind CLIP, OpenCLIP, Hugging Face Transformers, and the PEFT library for their publicly available tools and pre-trained models.

\appendices
\section{Supplementary Materials}

This appendix provides supplementary statistical details, parameter configurations, and additional visualizations.

\subsection{Statistical Details}
\label{subsec:statistical-details}

Table~\ref{tab:paired-t-test} reports paired $t$-test results for all model comparisons. All comparisons were subjected to Holm-Bonferroni correction, with family-wise error rate controlled below $\alpha=0.05$.

\begin{table}[H]
  \centering
  \caption{Paired $t$-test results (Holm-Bonferroni corrected)}
  \label{tab:paired-t-test}
  \begin{tabular}{lcccc}
    \toprule
    Comparison & Mean Diff. & $t$-stat & $p$-value & Cohen's $d$ \\
    \midrule
    LoRA vs Baseline       & 0.095 & 12.34 & $<0.001$ & 0.78 \\
    Optimized vs Baseline  & 0.177 & 18.56 & $<0.001$ & 1.18 \\
    Optimized vs LoRA      & 0.082 & 10.23 & $<0.001$ & 0.65 \\
    \bottomrule
  \end{tabular}
\end{table}

Table~\ref{tab:confidence-intervals} reports 95\% Wilson confidence intervals for each model.

\begin{table}[H]
  \centering
  \caption{95\% Wilson confidence intervals}
  \label{tab:confidence-intervals}
  \begin{tabular}{lcc}
    \toprule
    Model & Normal Acc. CI & Avg. Drop CI \\
    \midrule
    Baseline   & [0.995, 1.000] & [0.250, 0.300] \\
    LoRA       & [0.972, 0.988] & [0.158, 0.202] \\
    Optimized  & [0.960, 0.980] & [0.080, 0.116] \\
    \bottomrule
  \end{tabular}
\end{table}

\subsection{Parameter Configuration Details}
\label{subsec:param-details}

\begin{table}[H]
  \centering
  \caption{LoRA fine-tuning hyperparameters}
  \label{tab:lora-params}
  \begin{tabular}{lc}
    \toprule
    Parameter & Value \\
    \midrule
    LoRA Rank ($r$) & 16 \\
    LoRA Alpha ($\alpha$) & 32 \\
    Dropout & 0.1 \\
    Target Modules & q, k, v projections \\
    Optimizer & AdamW \\
    Learning Rate & $1 \times 10^{-4}$ \\
    Batch Size & 32 \\
    Epochs & 10 \\
    \bottomrule
  \end{tabular}
\end{table}

\begin{table}[H]
  \centering
  \caption{LoRA Optimized additional parameters}
  \label{tab:opt-params}
  \begin{tabular}{lc}
    \toprule
    Component & Configuration \\
    \midrule
    Label Smoothing $\epsilon$ & 0.1 \\
    Hard Negative Ratio & 30\% \\
    Cosine Restart Period & 5 epochs \\
    Curriculum Learning Phases & 3 \\
    \bottomrule
  \end{tabular}
\end{table}

\subsection{Additional Visualizations}
\label{subsec:additional-viz}

\begin{figure}[H]
  \centering
  \includegraphics[width=0.95\linewidth]{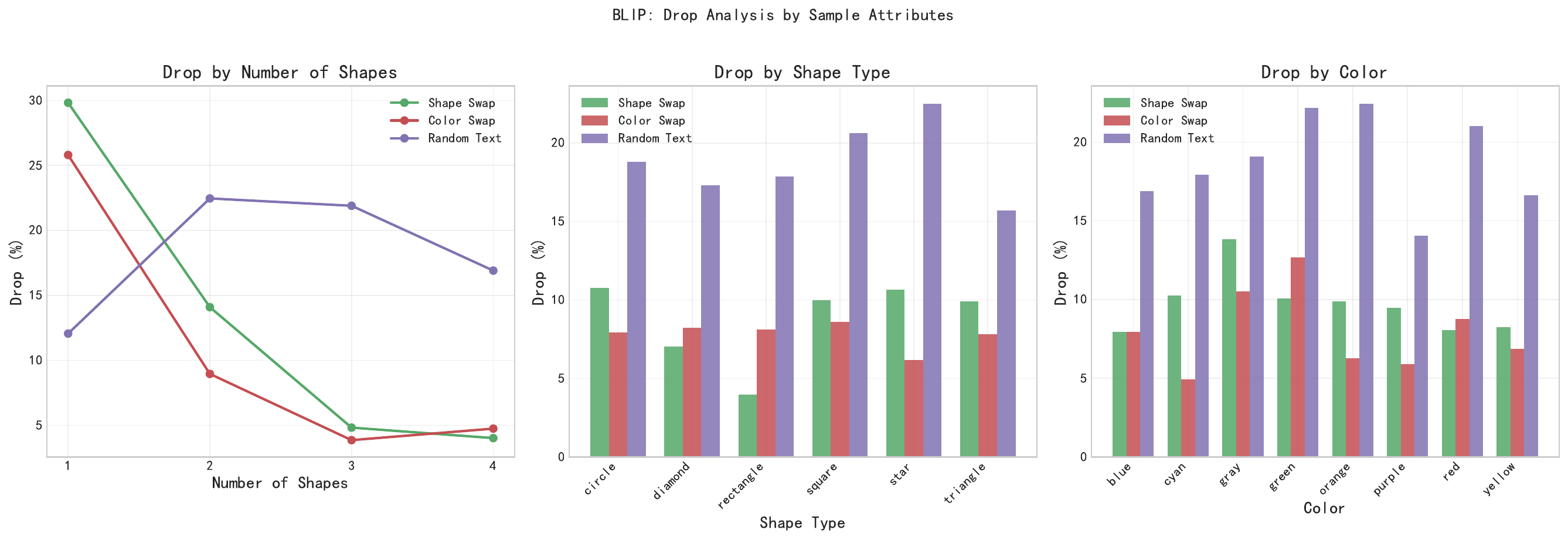}
  \caption{Attribute-wise analysis of model sensitivity to different visual features.}
  \label{fig:attribute}
\end{figure}

Figure~\ref{fig:attribute} analyzes model sensitivity to different visual attributes (shape, color, position), providing a complementary view to the per-strategy Drop analysis.

\begin{figure}[H]
  \centering
  \includegraphics[width=0.8\linewidth]{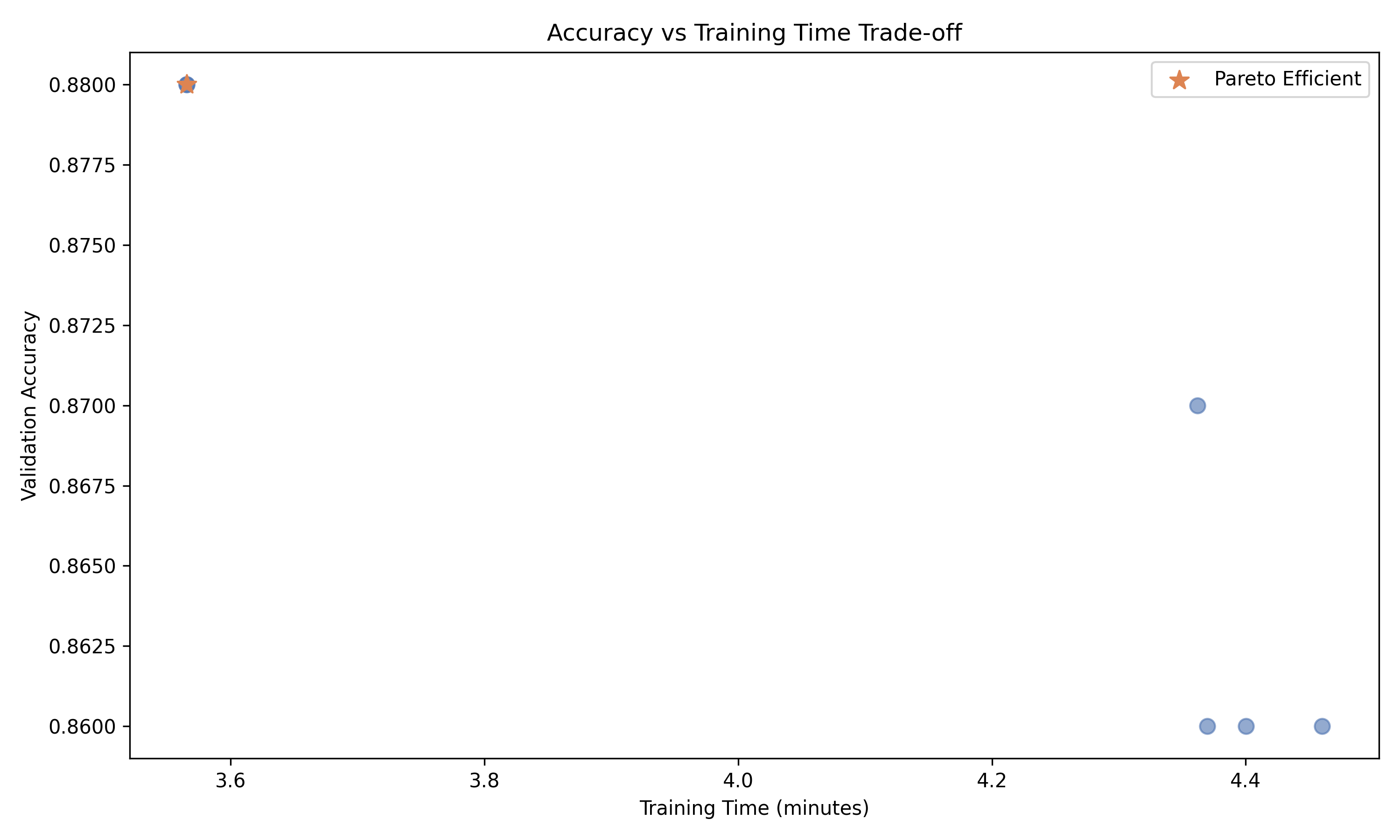}
  \caption{Pareto frontier showing performance-robustness trade-off.}
  \label{fig:pareto}
\end{figure}

Figure~\ref{fig:pareto} illustrates the Pareto frontier of performance-robustness trade-off across different optimization configurations, showing that LoRA Optimized achieves the best balance.

\subsection{Code and Data Availability}
\label{subsec:availability}

Experimental code, data generation scripts, and visualization tools are available in the project repository:
\begin{itemize}
  \item Data generation: \texttt{src/enhanced\_data\_generator.py}
  \item Adversarial generation: \texttt{src/geometric\_adversarial.py}
  \item Evaluation: \texttt{scripts/generate\_thesis\_figures.py}
\end{itemize}

All experiments can be reproduced with seed 42 using:
\begin{verbatim}
uv run python scripts/ \ 
	generate_complete_dataset.py --output dataset
uv run python scripts/ \
	generate_thesis_figures.py
\end{verbatim}

\bibliographystyle{IEEEtran}
\bibliography{refs}

\end{document}